# An Algorithm for Computing Probabilistic Propositions

Gregory F. Cooper

Medical Computer Science Group
Knowledge Systems Laboratory
Stanford University
Medical School Office Building, Room 217
Stanford, California 94305

## Abstract

An algorithm for computing probabilistic propositions is presented. It assumes the availability a single external routine for computing the probability of one instantiated variable[1], given a conjunction of other instantiated variables. Although the time complexity of the algorithm is exponential in the size of a query, it is polynomial in the size of a number of common types of queries.

## 1. Introduction

This paper presents an algorithm for computing the probability of a propositional logic sentence in the context of a set of known probabilities. Probability inference algorithms have typically been developed to calculate $P(S_1 \mid S_2)$, where $S_1$ is either a single instantiated variable or a conjunction of instantiated variables, and $S_2$ is a conjunction of instantiated variables [Cheeseman 83, Lemmer 83, Cooper 84, Shachter 86, Pearl 86, Pearl 87]. When $S_1$ is a single instantiated variable, we call these *single variable (SV) algorithms*.

We will extend probability queries to the case of $S_1$ and $S_2$ being well-formed formulas in propositional logic (propositions). Thus, it will be possible to apply disjunction, conjunction, and negation to variables in both the conditioning and conditioned part of a probability query. The propositional probability query algorithm for performing these type calculations, called PPQ, is a simple procedure based on calls to an SV algorithm. Thus, past implementations of SV algorithms can be easily augmented to answer more general propositional probability queries. Furthermore, PPQ can usually answer queries much more efficiently than a brute-force technique which explicitly sums over the entire joint probability space of the domain variables.

PPQ is a specialization of general probabilistic logic algorithms [Nilsson 86]. It handles only propositional logic rather than first order predicate logic. It also assumes that the background probabilistic information is sufficient for an SV algorithm to compute a unique (point) probability, rather than an upper and lower bound (interval) probability.

---

[1] The term *instantiated variable* is used to denote a variable with a known, assigned value.

## 2. The Principal Steps in the Algorithm

The main steps underlying PPQ involve handling conditional probabilities and the propositional relations AND ($\wedge$) and OR ($\vee$). These are shown below. A overbar will be used to designate negation.

**Step 1.** Since $P(S_1 \mid S_2)$ can be expressed as $P(S_1 \wedge S_2)/P(S_2)$, the task of computing conditional propositional probability queries is readily decomposed into computing two marginal propositional probabilities.

**Step 2.** The next step is to transform a marginal propositional probability $P(S)$ into a form $P(S')$ in which $S'$ contains only conjunctions and negations. This can be done by successive applications of de Morgan's law, namely, $X_1 \vee X_2 \Rightarrow \overline{\overline{X_1} \wedge \overline{X_2}}$.

**Step 3.** If $S'$ consists only of a conjunction of n instantiated variables of the form $X_1 \wedge X_2 \wedge ... \wedge X_n$, then by application of the chain-rule of conditional probabilities we know that:

$$P(S') = P(X_1 \mid X_2 \wedge ... \wedge X_n) \, P(X_2 \mid X_3 \wedge ... \wedge X_n) \, ... \, P(X_{n-1} \mid X_n) \, P(X_n)$$

Note that each of these terms can be computed by an SV algorithm.

**Step 4.** If $S'$ consists of complex variable groupings, then it is successively simplified. Such groupings are due to negation operators. The key to simplifying $S'$ is to remove negation operators. Suppose $S'$ equals $\overline{S'_1 \wedge S'_2} \wedge S'_3$, where each of these three terms is a proposition. In order to simplify this expression, the negation operator, which is grouping $S'_1$ and $S'_2$, can be removed as follows:

$$\begin{aligned}
P(S') &= P(\overline{S'_1 \wedge S'_2} \wedge S'_3) \\
&= P(\overline{S'_1 \wedge S'_2} \mid S'_3) \, P(S'_3) \\
&= (1 - P(S'_1 \wedge S'_2 \mid S'_3)) \, P(S'_3) \\
&= \left(1 - \frac{P(S'_1 \wedge S'_2 \wedge S'_3)}{P(S'_3)}\right) P(S'_3) \\
&= P(S'_3) - P(S'_1 \wedge S'_2 \wedge S'_3)
\end{aligned}$$

The terms $P(S'_3)$ and $P(S'_1 \wedge S'_2 \wedge S'_3)$ in the last line above are the only ones which must be computed; the derivation is just to clarify how the final line was obtained. Thus, the heart of Step 4 is the use of a simple elementary probability relation, namely,

381

$P(\overline{X} \wedge Y) = P(Y) - P(X \wedge Y)$, where X and Y are arbitrary propositions. In the above example, recursive application of this simplification rule to $P(S'_3)$ and $P(S'_1 \wedge S'_2 \wedge S'_3)$ will eventually yield probabilities with terms that consist of only a conjunction of instantiated variables. Step 3 can then be applied to determine the value of each of these probability terms and thus the value of $P(S')$. Note that none of the above four steps assumes that variables are binary. Thus, PPQ can answer queries that contain instantiated multi-valued variables.

## 3. An Example

As an example, consider a query to calculate $P(x_1 \vee x_2 \mid x_3 \vee \overline{x_4})$. The steps that PPQ takes in answering this query are shown below. Note that some probabilities occur more than once (e.g., $P(\overline{x_3} \wedge x_4)$), and caching would be useful in such cases.

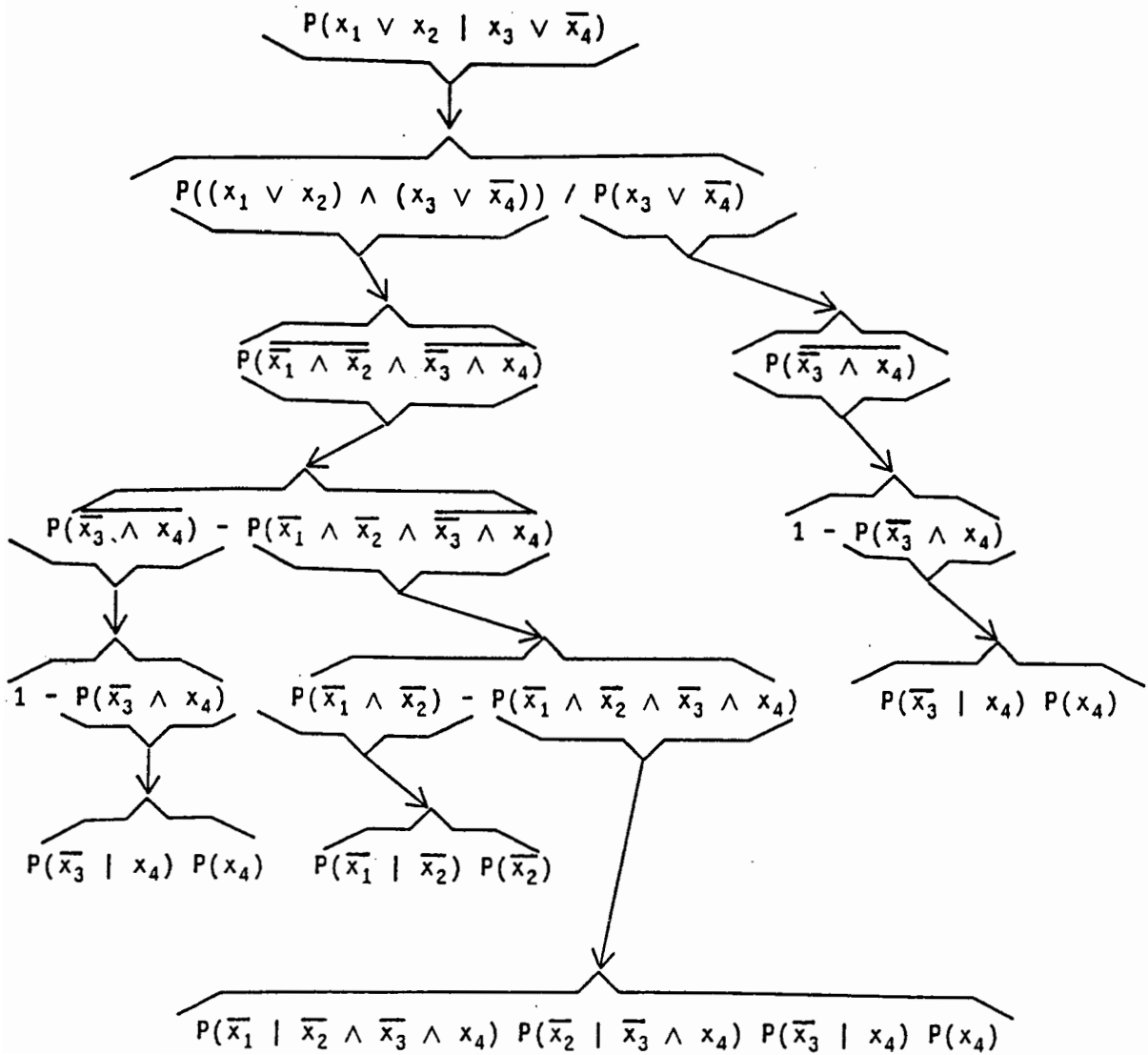

## 4. Time Complexity Analysis

The worst-case computational time complexity of PPQ is $O(m\, 2^m\, g(n))$, where m is the number of variables in a query, n is the total number of variables in the knowledge base, and $g(n)$ is the worst-case time complexity for a given SV algorithm to compute a probability using a knowledge base of size n. In the worst-case, the SV algorithm calculation, as reflected in $g(n)$, can be quite expensive. For example, using belief networks, this inference problem is known to be NP-hard [Cooper 87]. On the other hand, there are SV algorithms with an $O(n)$ time complexity for special belief network topologies, such as networks with only one pathway between any two nodes [Pearl 86]. However, the main issue in the analysis of PPQ is the time complexity incurred due to calculations other than those of the SV algorithm. This is reflected in the $m\, 2^m$ term above. Therefore, the remainder of our complexity analysis will focus on the derivation of $O(m\, 2^m)$ as the upper bound on the number of calls to an SV algorithm, for a query of size m.

The derivation of $O(m\, 2^m)$ as a worst-case upper bound is more easily understood in terms of the number of negation operators (produced in Step 2) that span two or more variables. Suppose there are q such negation operators for a given query. In Step 4, an inference problem of size q is reduced to two inference problems, each with size no greater than $q - 1$. Thus, the recurrence relation $F(q) = 2\, F(q - 1)$ is an upper bound on the time complexity of PPQ. $F(0)$ is upper bounded by m, since the absence of negation operators leads to an SV algorithm being called at most m times to compute a conjunction of at most m variables (Step 2 above). The solution to this recurrence relation is $F(q) = m\, 2^q$. The $O(m\, 2^m)$ upper bound result is obtained by substituting m for q, since $m > q$ for all queries that contain no redundant (unnecessary) negation operators. A query with redundant negation operators can be converted into a non-redundant query format in $O(m + q)$ time. Although $O(m\, 2^m)$ is an exponential worst-case upper bound, it is important to note that the $m\, 2^m$ calls to an SV algorithm are only exponential in the size of the *query*, not in the size of the entire knowledge base or probability space.

A lower bound on the number of SV algorithm calls that are required to compute a query of the form $P(\overline{X_1 \wedge X_2} \wedge \overline{X_3 \wedge X_4} \wedge ... \wedge \overline{X_{m-1} \wedge X_m})$ can be shown to be $\Omega(m\, 2^m)$. Thus, in the worst-case, the upper bound time complexity is the same as the lower bound time complexity. Therefore, the worst-case time complexity of PPQ is exponential, as a function of the query size.

Although the worst-case time complexity of PPQ may be intractable, there are a number of practical type queries which do not require an inordinate number of calls to an SV algorithm. For example, suppose $P(Y \mid X)$ is to be computed, for propositions Y and X. Furthermore, suppose that X is a conjunction of instantiated variables; in reality, this is often the case. Let m be the combined number of variable references in both Y

383

and X. The number of SV algorithm calls required to compute P(Y | X) will be O(m) when Y is of one of the following forms, and $Y_i$ and $Y_{ij}$ are single instantiated variables:

1. $Y_1 \wedge Y_2 \wedge ... \wedge Y_k$.
   This is a very common type of query. It is calculated by simply using Step 3 above. This form of query can be used to determine the probability of a multi-etiology diagnosis.

2. $Y_1 \vee Y_2 \vee ... \vee Y_k$.
   This type of query could be used to determine the probability that at least one etiology, from among a given set of etiologies, has occurred.

3. $(Y_{11} \wedge Y_{12} \wedge ... \wedge Y_{1s_1}) \vee (Y_{21} \wedge Y_{22} \wedge ... \wedge Y_{2s_2}) \vee ...$
   $\vee (Y_{r1} \wedge Y_{r2} \wedge ... \wedge Y_{rs_r})$,
   where r is bounded from above by some constant. This is a generalization of 1 and 2 above. This type query can be used to determine the probability that at least one multi-etiology hypothesis is true, from among a given set of multi-etiology hypotheses.

4. $(Y_{11} \vee Y_{12} \vee ... \vee Y_{1s_1}) \wedge (Y_{21} \vee Y_{22} \vee ... \vee Y_{2s_2}) \wedge ...$
   $\wedge (Y_{r1} \vee Y_{r2} \vee ... \vee Y_{rs_r})$,
   where r is bounded from above by some constant. This type query is also a generalization of 1 and 2 above. It can be used to determine the probability of a multi-etiology diagnostic hypothesis, where each etiology in the hypothesis is an abstract etiological class which is represented by disjunction.

Thus, although in the worst-case the time complexity of PPQ is exponential in the size of a query, it is polynomial in the size of a number of common types of queries. This pragmatic tractability, plus the generality and simplicity of PPQ, make it a potentially useful algorithm in systems that reason probabilistically.

### Acknowledgements
I wish to thank David Heckerman and Eric Horvitz for valuable discussions about this topic. This work has been supported by grant LM-07033 from the National Library of Medicine. Computer facilities were provided by the SUMEX-AIM resource under NIH grant RR-00785.